\pdfoutput=1

\documentclass[11pt]{article}

\usepackage{acl2023}
\usepackage{times}
\usepackage{latexsym}

\usepackage[T1]{fontenc}

\usepackage[utf8]{inputenc}

\usepackage{microtype}

\usepackage{graphicx}
\usepackage{multirow}
\usepackage{amsmath}
\usepackage{amsfonts}
\usepackage{bm}
\usepackage{booktabs}
\usepackage{subfigure}
\usepackage{verbatim}
\usepackage{amssymb}
\usepackage{enumitem}
\usepackage{algorithm}
\usepackage{algorithmic}
\usepackage{arydshln}
\usepackage{hyperref}
\usepackage{inconsolata}

\title{Uncertainty-Aware Unlikelihood Learning Improves Generative Aspect Sentiment Quad Prediction}

\author{Mengting Hu\textsuperscript{1} \quad Yinhao Bai\textsuperscript{1} \quad Yike Wu\textsuperscript{1}\thanks{\; Corresponding author.} \quad Zhen Zhang\textsuperscript{1} \\ {\bf Liqi Zhang\textsuperscript{2}} \quad {\bf Hang Gao\textsuperscript{3}} \quad {\bf Shiwan Zhao\textsuperscript{}\thanks{\; Independent researcher.}} \quad {\bf Minlie Huang\textsuperscript{3}} \\
\textsuperscript{1} Nankai University, \textsuperscript{2} Tiangong University, \textsuperscript{3} Tsinghua University \\
{\tt \{mthu, wuyike\}@nankai.edu.cn,} {\tt \{yinhao, zhangzhen\}@mail.nankai.edu.cn} \\
{\tt gaohang@mail.tsinghua.edu.cn,} {\tt aihuang@tsinghua.edu.cn}
}

\begin{document}
\maketitle
\begin{abstract}


Recently, aspect sentiment quad prediction has received widespread attention in the field of aspect-based sentiment analysis. Existing studies extract quadruplets via pre-trained generative language models to paraphrase the original sentence into a templated target sequence. However, previous works only focus on \emph{what to generate} but ignore \emph{what not to generate}. We argue that considering the negative samples also leads to potential benefits. In this work, we propose a template-agnostic method to control the token-level generation, which boosts original learning and reduces mistakes simultaneously. Specifically, we introduce Monte Carlo dropout to understand the built-in uncertainty of pre-trained language models, acquiring the noises and errors. We further propose marginalized unlikelihood learning to suppress the uncertainty-aware mistake tokens. Finally, we introduce minimization entropy to balance the effects of marginalized unlikelihood learning. Extensive experiments on four public datasets demonstrate the effectiveness of our approach on various generation templates\footnote{Experimental codes and data are available at: \url{https://github.com/byinhao/UAUL}.}.


\end{abstract}

\section{Introduction}
Recently, aspect sentiment quad prediction (ASQP) has received extensive attention in the field of aspect-level sentiment analysis. ASQP targets a comprehensive sentiment understanding and extracts four elements of aspect sentiment, including 1) \emph{aspect term (at)} which is the concrete aspect description; 2) \emph{opinion term (ot)} suggesting the specific opinion expression towards the aspect; 3) \emph{aspect category (ac)} denoting the aspect class; 4) \emph{sentiment polarity (sp)} indicating the sentiment class of the aspect. For example, given a comment sentence \emph{``Service was good and food was wonderful''}, ASQP aims to recognize two quadruples (\emph{Service}, \emph{good}, \emph{service general}, \emph{positive}) and (\emph{food}, \emph{wonderful}, \emph{food quality}, \emph{positive}).



Existing works have pointed out two promising research directions. \citet{cai2021aspect} propose a pipeline-based method, using the properties of four elements and designing first-extract-then-classify two stages. Another direction leverages generation-based pre-trained language models. ASQP is addressed in an end-to-end manner by \emph{``re-writing''} a sentence into a structured target sequence \cite{zhang2021towards-generative,zhang2021aspect-sentiment,hu2022improving}. With pre-defined templates, quadruples can be easily decoded from the target sequence. Due to its simplicity and effects, the second paradigm gradually becomes the main streaming in ASQP \cite{hu2022improving}.
\begin{figure}[t]
\centering
\includegraphics[width=0.48\textwidth]{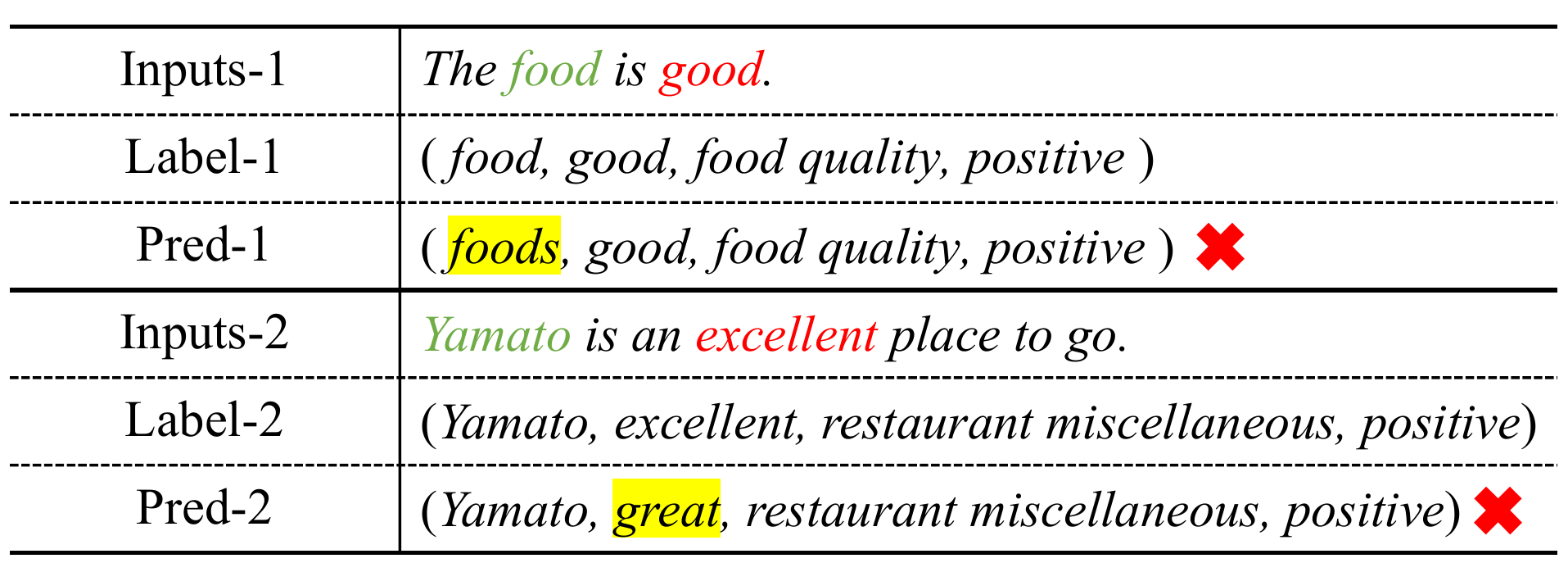} 
\caption{Two predicted error cases are shown. Pred denotes the prediction. The results of Label and Pred are shown in the order of ($at$, $ot$, $ac$, $sp$), and the highlighted parts are the predicted error items.}
\label{fig:error}
\end{figure}


However, no matter designing good templates \cite{zhang2021aspect-sentiment,bao2022aspect} or using data augmentation \cite{hu2022improving}, previous generation-based works only focus on \emph{what to generate} but ignore \emph{what not to generate}. Learning signals of negative effects are also crucial for accurate extraction. The reason is that ASQP is not a typical text-generation task, such as dialog \cite{liu2021towards} or storytelling \cite{xu2020megatron}. Semantic-similar or ambiguous words are harmful for extraction. Two failed cases of pre-trained language models are presented in Figure \ref{fig:error}. In the first case, the aspect term \emph{``food''} is easily confused with \emph{``foods''}. And the second case also implies that the opinion term \emph{``excellent''} can be wrongly decoded as \emph{``great''}. Though these words do not obviously change the semantics, they lead to complete mistakes for ASQP. Therefore, how to make language models to avoid errors motivates us. 


In this paper, we propose uncertainty-aware unlikelihood learning (UAUL) to guide the likelihood learning (\emph{what can be generated}) and marginalized unlikelihood learning (\emph{what not to generate}) simultaneously. Concretely, \emph{what to generate} is in light of the sequence-to-sequence learning objective. Target sequences are constructed with pre-defined templates, providing semantic and syntactic structured information. As for \emph{what not to generate}, we argue that the noise and errors present in the pre-trained model are due to the uncertainty of the model itself. Therefore, we introduce the Monte Carlo dropout (MC dropout) \cite{gal2016dropout} to obtain built-in negative samples of pre-trained models. By dropping out random parameters of the last layer followed by the decoder, i.e. language model head, multiple predictions can be attained, which further tell the inherent errors of language models.



Moreover, with uncertainty-aware negative samples, we further propose marginalized unlikelihood learning (MUL) to suppress the probability of them. The marginalization could promote the gap between correct and error tokens, making models to better distinguish semantically similar or ambiguous words. Finally, MUL reduces the probability of noises. This might enlarge the probability of other errors, since the vocabulary set of language models is with the scale. Hence, to balance the influences of MUL, we propose to minimize the entropy of uncertainty-aware probability distributions.

In summary, the contributions of this paper are as follows:

\noindent
\begin{itemize}[leftmargin=*]
    \item We study generative ASQP task from the view of \emph{what not to generate}. To the best of our knowledge, this is the first work to study negative samples in this task. We propose uncertainty-aware unlikelihood learning to avoid the intrinsic mistakes of pre-trained language models. 
    \item  Specifically, the model uncertainty is comprehended with MC dropout. And the built-in errors are reduced with the proposed marginalized unlikelihood learning and minimization entropy. Our method is template-agnostic and can be easily applied to various target templates.
    
    \item Experimental results on four public datasets Rest15, Rest16, Restaurant, and Laptop demonstrate that UAUL has universal effectiveness  on various templates.
    
\end{itemize}

    
\section{Methodology}

\subsection{Formulation and Overview}

Given a sentence $\bm{x}$, aspect sentiment quad prediction (ASQP) aims to predict all aspect-level quadruplets $\{({at},{ot},{ac},{sp})\}$. Following the previous generation-based works \cite{zhang2021aspect-sentiment,hu2022improving}, we define projection functions to map the quadruplets $({at},{ot},{ac},{sp})$ into semantic values $(x_{at},x_{ot},x_{ac},x_{sp})$. Concretely, 1) if aspect term $at$ is explicit, $x_{at}={at}$, otherwise $x_{at}=$\emph{``it''}; 2) if opinion term $ot$ are explicitly mentioned, $x_{ot}=ot$, otherwise it is mapped as \emph{``NULL''} if being implicitly expressed; 3) aspect category $ac$ is transformed into words, such as $x_{ac}=$\emph{``food quality''} for $ac=$\emph{``food\#quality''}; 4) the sentiment polarity $sp\in$ \{\emph{positive}, \emph{neutral}, \emph{negative}\}, is mapped into words with sentiment semantics \{\emph{great}, \emph{ok}, \emph{bad}\}, respectively.



Based on the above rules, the values are fed into a template $\mathcal{T}$ to form the target sequence. For instance, a template follows the cause and effect semantic relationship ``$x_{ac} \text{ is } x_{sp} \text{ because } x_{at} \text{ is } x_{ot}$'' \cite{zhang2021aspect-sentiment} or uses special markers ``$\mathtt{[AT]}\ x_{at}\ \mathtt{[OT]}\ x_{ot}\ \mathtt{[AC]}\ x_{ac}\ \mathtt{[SP]}\ x_{sp}$'' \cite{hu2022improving}. If a sentence contains multiple quadruplets, the templated sequences are concatenated with a special marker $\mathtt{[SSEP]}$ to obtain the final target sequence $\bm{y}$.

    \begin{figure*}[t]
    \centering
    \includegraphics[width=0.85\textwidth]{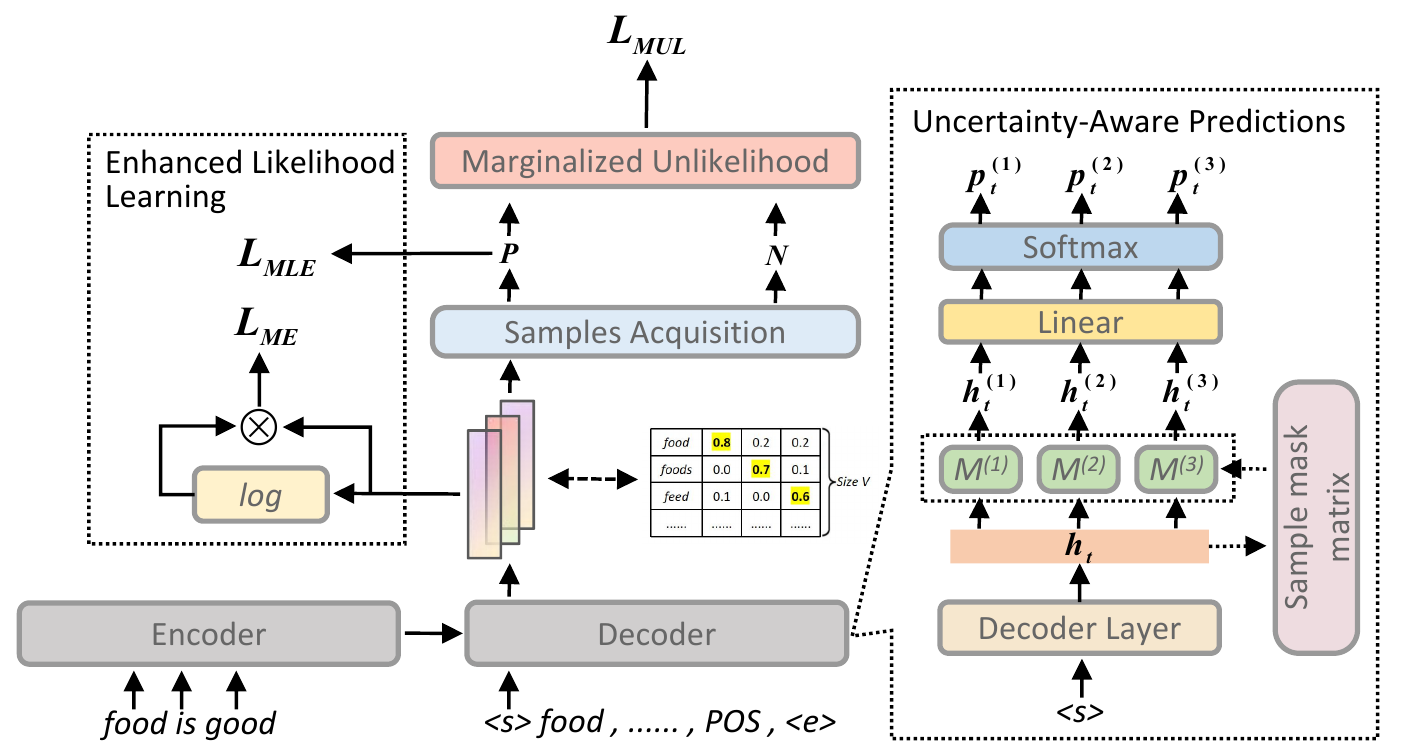}
    \caption{An overview of the proposed uncertainty-aware unlikelihood learning (UAUL). We present the details via an example of the first decoding time step. The beginning token \emph{``<s>''} yields its next token, i.e. \emph{``food''}, where the output is enhanced as three uncertainty-aware probability distributions $\{\bm{p_t^{(i)}}\}$. The largest probability is highlighted and chosen as a negative sample.}
    \label{fig:overview}
    \end{figure*}



As shown in Figure \ref{fig:overview}, an input sentence is first fed into the encoder-decoder framework. We exploit the pre-trained language model T5 \cite{JMLR:v21:20-074}. To deal with negative samples\footnote{It is worth noting that negative samples indicate the noisy tokens in the vocabulary set rather than a sentence.}, we first acquire multiple uncertain-aware samples via MC dropout for each decoding time step. Then these samples are fed to calculate marginalized unlikelihood learning loss. Finally, to enhance the learning of target sequence and balance the effects of MUL, we design enhanced likelihood learning for uncertainty-aware samples. Next, we will introduce the components in detail. %


\subsection{Uncertainty-Aware Samples Acquisition}
    

As depicted in Figure \ref{fig:error}, semantic-similar or ambiguous words lead to complete error predictions for ASQP task. The in-depth reason relies on that language models are pre-trained based on distributional semantics theory \cite{boleda2020distributional}, producing alike representations for words that frequently appear in similar contexts, such as \emph{``excellent''} and \emph{``great''}. Then when extracting aspect quadruplets, language models are not sure which one is more accurate. Understanding the inherent uncertainty of language models may have potential benefits. To achieve this goal, we re-design the decoder of T5 and adopt MC dropout \cite{gal2016dropout} to obtain valuable samples. 




\noindent
\textbf{Uncertainty-Aware Predictions} \; The target sequence $\bm{y}$ is fed into the decoder as teacher forcing \cite{williams1989learning} during training. The decoder's inner layers are depicted in the right plot of Figure \ref{fig:overview}. Here, we use the beginning token \emph{``<s>''} as an example to illustrate the details of each time step. We obtain a representation for each token with multiple transformer-based self-attention mechanisms of the decoder layer. 
\begin{equation}
\bm{h_t}=\mathrm{Enc-DecLayer}(\bm{x}, \bm{y_{<t}})
    \label{eq:encoder-decoder}
\end{equation}
where $\bm{h_t}$ is calculated based on the input sequence $\bm{x}$ and previous outputs $\bm{y_{<t}}$. $\mathrm{Enc-DecLayer}$ indicates the encoder module and the decoder layer.
    
Then, following \newcite{vazhentsev2022uncertainty}, we only exploit the last dropout layer, which is much less computationally expensive than the standard MC dropout. Specifically, an uncertain-aware representation is obtained by sampling a random mask matrix $M^{(i)}$.
    \begin{equation}
    \bm{h_t^{(i)}}=M^{(i)}\times\bm{h_t}
        \label{eq:get-masks}
    \end{equation}
where sampling $M^{(i)}$ follows the Bernoulli distribution $Bernoulli(1-p)$ and $p\in[0, 1]$ is the dropout rate. Then the output is calculated based on the uncertainty-aware representations.
    \begin{equation}
    \bm{p_t^{(i)}}=\mathrm{softmax}(W^{\mathrm{T}}\bm{h_t^{(i)}})
        \label{eq:mc dropout}
    \end{equation}
where $W$ maps $\bm{h_t^{(i)}}$ into a vector and $\bm{p_t^{(i)}}$ indicates the probability distribution over the vocabulary set. We dropout multiple times and attain multiple output distributions at $t$-th time step $\{\bm{p_t^{(i)}}\}$ and $i\in[1,K]$. $K$ is the number of MC forward computations.

\begin{algorithm}[t]
\caption{Samples Acquisition}
    \begin{algorithmic}
        \STATE \textbf{Input}: Output probability distributions $\{\bm{p_t^{(i)}}\}$, ground-truth $y_t$
        \STATE \textbf{Define}: $P_t=\varnothing$,  $N_t=\varnothing$
        \FOR {$i = 1,2,...,K$}
            \STATE $c\gets \mathrm{argmax}(\bm{p_t^{(i)}})$
            \IF {$c$ !$=$ $y_t$}
                \STATE $N_t \gets N_t \cup \bm{p_t^{(i)}}[c]$
            \ENDIF
        \STATE $P_t \gets P_t \cup \bm{p_t^{(i)}}[y_t]$
        \ENDFOR
    \RETURN $P_t, N_t$
    \end{algorithmic}
\label{al:sample negative}
\end{algorithm}

\noindent
\textbf{Samples Acquisition} \; Based on multiple uncertain-aware output probability distributions, we then acquire key samples as described in Algorithm \ref{al:sample negative}. Note that this algorithm displays the acquisition of samples for time step $t$. For the positive samples, we mainly concentrate on the probability of the ground-truth token $y_t$ out of each distribution. For the negative samples, we choose the largest wrong prediction. In this way, both the positive and negative samples are integrated with the uncertainty of language models (i.e. various probabilities). Meanwhile, negative samples are also difficultly distinguishable ones. 


\subsection{Marginalized Unlikelihood Learning}  

Then with these chosen key samples, we propose marginalized unlikelihood learning to explicitly control their optimization. As an example shown in Figure \ref{fig:overview}, we have three output distribution $\{\bm{p_t^{(1)}}, \bm{p_t^{(2)}}, \bm{p_t^{(3)}}\}$. The highlighted probabilities are the largest in that distribution. With Algorithm \ref{al:sample negative}, we obtain $P_t=\{0.8, 0.2, 0.2\}$ and $N_t=\{0.7, 0.6\}$, where $P_t$ and $N_t$ are sampled positive and negative samples, respectively. These probabilities are further utilized to calculate in Eq. (\ref{eq:MUL}). 

    
\begin{footnotesize}
    \begin{equation}
        \mathcal{L}_{MUL}=\sum_{t=1}^{n}\mathrm{log}[1+\sum_{k=1}^{|P_t|}\sum_{l=1}^{|N_t|}\exp(\alpha(N_t^l-P_t^k+m))]
        \label{eq:MUL}
    \end{equation}
    
\end{footnotesize}
where $\alpha$ is the scale hyperparameter. $m$ is the margin between positive and negative samples. $n$ is the length of the target sequence. $|P_t|$ means the number of samples in $P_t$. 

It is worth noting that the proposed MUL is based on the largest probability in every uncertainty-aware distribution. Putting it to $N_t$ according to whether it is correct. The reason is that softmax probabilities tend to be overconfident \cite{pmlr-v70-guo17a}, making all other probabilities very small except for the largest one. Then our method can better select easily-mistaken samples from multiple distributions. 


\subsection{Enhanced Likelihood Learning}
Except for dealing with the noise issue, \emph{what to generated} is still important to obtain task-specific semantic and structured knowledge. We exploit the original likelihood training to optimize the positive sample probabilities on multiple uncertainty-aware probability distributions.
\begin{equation}
    \mathcal{L}_{MLE}=-\frac{1}{K}\sum_{i=1}^K\sum_{t=1}^n\bm{y_t}\mathrm{log}\bm{p_t^{(i)}}
    \label{eq:likelihood}
\end{equation}
where $\bm{y_t}$ denotes a ground true one-hot vector at the time step $t$.

Though MUL reduces the probability of noise, it might enlarge the probability of other errors. Take Figure \ref{fig:overview} as an example, reducing the probability of \emph{``foods''} may disperse the numerical to other noisy words, such as \emph{``feed''} or \emph{``apple''}. Thus, the optimization of MUL and likelihood is not fully consistent, which in turn affects the likelihood of learning. To balance likelihood and MUL, we introduce the minimization entropy (ME) loss term.
\begin{equation}
    \mathcal{L}_{ME}=-\sum_{i=1}^K\sum_{t=1}^{n} \bm{p_t^{(i)}}\mathrm{log}\bm{p_t^{(i)}}
    \label{eq:ME}
\end{equation}

By minimizing Eq. (\ref{eq:ME}), $\bm{p_t^{(i)}}$ will become more peak, that is to say, suppressing the noises simultaneously. In this way, our approach seeks a balance between MUL and likelihood, ensuring both accurate extraction and negative sample reduction.

\subsection{Joint Training Objective}
    
The final training objective is jointly to combine the above three losses.
        \begin{equation}
            \mathcal{L}=\mathcal{L}_{MLE}+\mathcal{L}_{MUL}+\mathcal{L}_{ME}
            \label{eq:MI}
        \end{equation}
    
\section{Experiments}

\begin{table}[]
\small
    \centering
    
    \scalebox{0.95}{
    \begin{tabular}{l cc cc cc}
    \toprule
     \multirow{2}{*}{Datasets} & \multicolumn{2}{c}{Train} & \multicolumn{2}{c}{Dev} & \multicolumn{2}{c}{Test} \\
     & \#S & \#Q & \#S & \#Q & \#S & \#Q \\
    \midrule
    $\mathtt{Rest15}$ & 834 & 1354 & 209 & 347 & 537 & 795 \\
    $\mathtt{Rest16}$ & 1264 & 1989 & 316 & 507 & 544 & 799 \\
    $\mathtt{Restaurant}$ & 2934 & 4172 & 326 & 440 & 816 & 1161 \\
    $\mathtt{Laptop}$ & 1530 & 2484 & 171 & 261 & 583 & 916 \\
\bottomrule
    \end{tabular}}
    \caption{Data statistics. \#S and \#Q denote the number of sentences and quadruplets respectively.}
    \label{table:data}
\end{table}

\subsection{Datasets}

To evaluate the proposed approach, we choose four publicly available datasets. $\mathtt{Rest15}$ and $\mathtt{Rest16}$ datasets are proposed by \citet{zhang2021aspect-sentiment}. They are based on previous SemEval tasks \cite{pontiki2015semeval,pontiki2016semeval}, and expanded with quadruplet annotations. \newcite{cai2021aspect} propose $\mathtt{Restaurant}$ and $\mathtt{Laptop}$ datasets. The $\mathtt{Restaurant}$ dataset is constructed based on the SemEval 2016 Restaurant dataset \cite{pontiki2016semeval} and its expansion datasets \cite{fan2019target,xu2020position}. The $\mathtt{Laptop}$ dataset is annotated based on the data collected on Amazon between 2017 and 2018. The statistics of datasets are displayed in Table \ref{table:data}.

\begin{table*}[!t]
\small
    \centering
    
    \begin{tabular}{l ccc ccc}
    \toprule
    \multirow{2}{*}{Methods} & \multicolumn{3}{c}{$\mathtt{Rest15}$} & \multicolumn{3}{c}{$\mathtt{Rest16}$} \\
    & $\mathtt{Pre}$ & $\mathtt{Rec}$ & $\mathtt{F1}$ & $\mathtt{Pre}$ & $\mathtt{Rec}$ & $\mathtt{F1}$ \\
    \midrule
    HGCN-BERT+BERT-Linear$^*$ \cite{zhang2021aspect-sentiment} & 24.43 & 20.25 & 22.15 & 25.36 & 24.03 & 24.68 \\
    HGCN-BERT+BERT-TFM$^*$ \cite{zhang2021aspect-sentiment} & 25.55 & 22.01 & 23.65 & 27.40 & 26.41 & 26.90 \\
    TASO-BERT-Linear$^*$ \cite{zhang2021aspect-sentiment} & 41.86 & 26.50 & 32.46 & 49.73 & 40.70 & 44.77 \\
    TASO-BERT-CRF$^*$ \cite{zhang2021aspect-sentiment} & 44.24 & 28.66 & 34.78 & 48.65 & 39.68 & 43.71 \\
    Extract-Classify-ACOS$^*$ \cite{cai2021aspect} & 35.64 & 37.25 & 36.42 & 38.40 & 50.93 & 43.77 \\
    \midrule
    GAS$^*$ \cite{zhang2021towards-generative} & 45.31 & 46.70 & 45.98 & 54.54 & 57.62 & 56.04 \\
    \quad +UAUL & \textbf{46.39} & \textbf{47.82} & \textbf{47.10} & \textbf{55.95} & \textbf{58.30} & \textbf{57.10} \\
    \hdashline
    Paraphrase$^*$ \cite{zhang2021aspect-sentiment} & 46.16 & 47.72 & 46.93 & 56.63 & 59.30 & 57.93 \\
    \quad +UAUL & \textbf{48.96} & \textbf{49.81} & \textbf{49.38} & \textbf{58.28} & \textbf{60.58} & \textbf{59.40} \\
    \hdashline
    Special\_Symbols$^*$ \cite{hu2022improving} & 48.24 & 48.93 & 48.58 & 58.74 & 60.35 & 59.53 \\
    \quad +UAUL & \textbf{49.12} & \textbf{50.39} & \textbf{49.75} & \textbf{59.24} & \textbf{61.75} & \textbf{60.47} \\
    \hdashline
    DLO$^*$ \cite{hu2022improving} & 47.08 & 49.33 & 48.18 & 57.92 & 61.80 & 59.79 \\
    \quad +UAUL & \textbf{48.03} & \textbf{50.54} & \textbf{49.26} & \textbf{59.02} & \textbf{62.05} & \textbf{60.50} \\
    \hdashline
    ILO$^*$ \cite{hu2022improving} & 47.78 & 50.38 & 49.05 & 57.58 & 61.17 & 59.32 \\
    \quad +UAUL & 46.84 & 49.53 & 48.15 & \textbf{58.23} & \textbf{61.35} & \textbf{59.75} \\
    \bottomrule
    \end{tabular}
    \caption{Evaluation results compared with baseline methods in terms of precision ($\mathtt{Pre}$, \%), recall ($\mathtt{Rec}$, \%) and F1 score ($\mathtt{F1}$, \%). The results of baseline methods, marked with $^*$, are obtained from this work \cite{hu2022improving}.}
    \label{table:results1}
    
\end{table*}

\subsection{Compared Methods}
We choose the following strong baseline methods and divided them into two types: i.e. \emph{non-generation} and \emph{generation}.

\noindent
\textbf{Non-Generation Baselines:} Traditional paradigm designs various stages to extract individual information separately.
\begin{itemize}[leftmargin=*]
    \vspace{-6pt}
    \item \textbf{Double-Propagation} \cite{10.1162/coli_a_00034} \; It is a classical method for triple extraction. \citet{cai2021aspect} adapt it to ASQP. All $\{at, ot, sp\}$ triplets are first extracted using double propagation, and then each triplet is assigned $ac$ to attain quad.
    \vspace{-6pt}
    \item \textbf{JET} \cite{xu2020position} \; It is an end-to-end framework for detecting triplet. \citet{cai2021aspect} first obtain $\{at, ot, sp\}$ triples with JET and then leveraged BERT to obtain $ac$.
    \vspace{-6pt}
    \item \textbf{HGCN-BERT+BERT} \cite{zhang2021aspect-sentiment} \; It is designed for learning syntactic dependencies for ASQP. Its variants include HGCN-BERT+BERT-Linear and HGCN-BERT+BERT-TFM according to the last layer.
    \vspace{-6pt}
    \item \textbf{TAS-BERT} \cite{wan2020target} \; It recognize triplets $\{at, ac, sp\}$ via learning dependencies. \citet{cai2021aspect} reformulate TAS-BERT to filter out invalid $\{ac, at\}$ pairs to get the final quadruplet. \citet{zhang2021aspect-sentiment} extend TAS-BERT to detect $ot$ for ASQP task. The variants are TASO-BERT-Linear and TASO-BERT-CRF.
    \vspace{-6pt}
    \item \textbf{Extract-Classify-ACOS} \cite{cai2021aspect} \; It first extracts aspect-opinion and then classifies category and sentiment, yielding the final quad.
\end{itemize}

 \begin{table*}[!t]
\small
    \centering
    
    \begin{tabular}{l ccc ccc}
    \toprule
    \multirow{2}{*}{Methods} & \multicolumn{3}{c}{$\mathtt{Restaurant}$} & \multicolumn{3}{c}{$\mathtt{Laptop}$} \\
    & $\mathtt{Pre}$ & $\mathtt{Rec}$ & $\mathtt{F1}$ & $\mathtt{Pre}$ & $\mathtt{Rec}$ & $\mathtt{F1}$ \\
    \midrule
    DP$^*$ \cite{cai2021aspect} & 34.67 & 15.08 & 21.04 & 13.04 & 00.57 & 08.00 \\
    JET$^*$ \cite{cai2021aspect} & 59.81 & 28.94 & 39.01 & 44.52 & 16.25 & 23.81 \\
    TAS-BERT$^*$ \cite{cai2021aspect} & 26.29 & 46.29 & 33.53 & 47.15 & 19.22 & 27.31 \\
    Extract-Classify-ACOS$^*$ \cite{cai2021aspect} & 38.54 & 52.96 & 44.61 & 45.56 & 29.48 & 35.80 \\
    \midrule
    GAS \cite{zhang2021towards-generative} & 57.09 & 57.51 & 57.30 & 43.45 & 43.29 & 43.37 \\
    \quad +UAUL & \textbf{58.69} & \textbf{59.26} & \textbf{58.97} & \textbf{43.58} & 42.98 & 43.28 \\
    \hdashline
    Paraphrase \cite{zhang2021aspect-sentiment} & 59.85 & 59.88 & 59.87 & 43.44 & 42.56 & 43.00 \\
    \quad +UAUL & \textbf{60.39} & \textbf{60.04} & \textbf{60.21} & \textbf{44.91} & \textbf{44.01} & \textbf{44.45} \\
    \hdashline
    Special\_Symbols \cite{hu2022improving} & 59.98 & 58.40 & 59.18 & 43.58 & 42.72 & 43.15 \\
    \quad +UAUL & \textbf{61.22} & \textbf{59.87} & \textbf{60.53} & \textbf{44.38} & \textbf{43.65} & \textbf{44.01} \\
    \hdashline
    DLO \cite{hu2022improving} & 60.02 & 59.84 & 59.93 & 43.40 & 43.80 & 43.60 \\
    \quad +UAUL & \textbf{61.03} & \textbf{60.55} & \textbf{60.78} & \textbf{43.78} & 43.53 & \textbf{43.65} \\
    \hdashline
    ILO \cite{hu2022improving} & 58.43 & 58.95 & 58.69 & 44.14 & 44.56 & 44.35 \\
    \quad +UAUL & \textbf{59.46} & \textbf{59.12} & \textbf{59.29} & 43.92 & 43.46 & 43.69 \\
    \bottomrule
    \end{tabular}
    \caption{Evaluation results compared with baseline methods in terms of precision ($\mathtt{Pre}$, \%), recall ($\mathtt{Rec}$, \%) and F1 score ($\mathtt{F1}$, \%). The results of baseline methods, marked with $^*$, are obtained from this work \cite{cai2021aspect}.}
    \label{table:results2}
    
\end{table*}

\noindent
\textbf{Generation Baselines:} Aspect sentiment quadruplets are fed into semantic templates to obtain a target sequence for generation learning. 
\begin{itemize}[leftmargin=*]
    \vspace{-6pt}
    \item \textbf{GAS} \cite{zhang2021towards-generative} \; It is the first work to reformulate all ABSA tasks as generation problems, and process all sub-tasks in a unified generation framework.
    
    \vspace{-6pt}
    \item \textbf{Paraphrase} \cite{zhang2021aspect-sentiment} \; It transforms the quadruplet extraction into a paraphrase generation through a predefined template.
    
    \vspace{-6pt}
    \item \textbf{Special\_Symbols} \cite{hu2022improving}\; It distinguishes the type of element in each position by special symbols.
    
    \vspace{-6pt}
    \item \textbf{DLO} \cite{hu2022improving} \; It designs dataset-level data augmentation via template-order permutation. The templates use special symbols.
    \vspace{-6pt}
    \item \textbf{ILO} \cite{hu2022improving} \; ILO designs data augmentation for each instance to find the good template order. The templates adopt special symbols.
\end{itemize}




\subsection{Experimental Results}

\subsubsection{Overall Results}

Experimental results are reported in Table \ref{table:results1} and Table \ref{table:results2}. Firstly, it can be observed that our method is effective in almost all experimental settings. Especially, compared with a strong baseline Paraphrase, Paraphrase+UAUL gains absolute F1 score improvements by +2.45\% (+5.22\% relatively), +1.47\% (+2.54\% relatively), +0.34\% (+0.57\% relatively), and +1.45\% (+3.37\% relatively) in $\mathtt{Rest15}$, $\mathtt{Rest16}$, $\mathtt{Restaurant}$ and $\mathtt{Laptop}$ datasets, respectively. Similarly, Special\_Symbols+UAUL achieves consistent improvements on all datasets, performing the best on $\mathtt{Rest15}$ dataset. Compared with DLO, DLO+UAUL also performs consistently better, achieving the best F1 scores of 60.50\% and 60.78\% on the $\mathtt{Rest16}$ and $\mathtt{Restaurant}$ datasets, respectively. These results demonstrate that the proposed UAUL can be easily applied to various templates with universal effectiveness. 

Moreover, we also see a few exceptions. For example, on $\mathtt{Laptop}$ dataset, UAUL causes the performances of GAS and ILO slightly decline. A possible reason is that $\mathtt{Laptop}$ dataset has a larger proportion of implicit information. Template treats implicit aspect term as \emph{``it''} and implicit opinion term as \emph{``NULL''}. Such implicit information makes it hard to understand quadruplets accurately. 

\tabcolsep=0.5cm
\begin{table}[]
\renewcommand\arraystretch{1.3}
\small
    \centering
    
    \setlength{\tabcolsep}{1mm}{
    \begin{tabular}{l|ccc|ccc}
    \hline
    Ratio & SS & +UAUL & $\Delta$ & Para & +UAUL & $\Delta$ \\
    \hline
    10\% & 29.05 & \textbf{31.48} & 2.43 & 26.68 & \textbf{30.55} & 3.87 \\
    15\% & 30.92 & \textbf{34.39} & 3.47 & 28.74 & \textbf{33.63} & 4.89 \\
    20\% & 36.83 & \textbf{37.07} & 0.24 & 33.60 & \textbf{36.70} & 3.10 \\
    25\% & 37.76 & \textbf{39.20} & 1.44 & 35.01 & \textbf{37.71} & 2.70 \\
    30\% & 39.41 & \textbf{41.03} & 1.62 & 37.12 & \textbf{40.25} & 3.13 \\
    35\% & 41.83 & \textbf{42.52} & 0.69 & 38.10 & \textbf{42.30} & 4.20 \\
    40\% & 42.46 & \textbf{44.17} & 1.71 & 39.48 & \textbf{44.35} & 4.87 \\
    45\% & 42.84 & \textbf{44.86} & 2.02 & 40.49 & \textbf{43.28} & 2.79 \\
    50\% & 44.64 & \textbf{45.69} & 1.05 & 42.48 & \textbf{45.30} & 2.82 \\
    \hline
    \end{tabular}}
    \caption{Evaluation results of low-resource scenario in terms of F1 (\%). Radio indicates the proportion of $\mathtt{Rest15}$ dataset's training data. SS and Para are Special\_Symbols and Paraphrase methods for short. $\Delta$ denotes the absolute improvements.}
    \label{table:Low-Resource}
    
\end{table}

\subsubsection{Low-Resource Scenario}


To further explore the performance of our proposed method in a low-resource environment, we train the model only with subsets of $\mathtt{Rest15}$. The results are reported in Table \ref{table:Low-Resource}. We can see that for both baseline methods, i.e. Special\_Symbols and Paraphrase, UAUL can bring consistent improvements with various data scales. In particular, with only 15\% training data, UAUL improves Special\_Symbols and Paraphrase significantly by +3.47\% (+11.22\% relatively) and +4.89\% (+17.01\% relatively) on F1 score, respectively. This verifies that UAUL shows more significant effectiveness in low-resource scenarios. A rational explanation is that low-resource might boost the overfitting of language models to the small-scale data. Then mistakes will occur more frequently, which are potentially distributed within the model and are caused by the uncertainty of the model itself. Our method helps the models to understand these potential errors well and addresses them to some extent. Therefore, UAUL is not only template-agnostic but also resource-friendly. 



\begin{table}[]
\small
    \centering
    
    \scalebox{0.95}{\setlength{\tabcolsep}{2.0mm}{
    \begin{tabular}{l ll cc cc}
    \toprule
    Datasets & Model & $\mathtt{Pre}$ & $\mathtt{Rec}$ & $\mathtt{F1}$ \\
    \midrule
    \multirow{6}{*}{$\mathtt{Rest15}$} & \textbf{Our} & \textbf{49.12} & \textbf{50.39} & \textbf{49.75} \\
    \hdashline
    & -ME & 49.11 & 50.18 & 49.64 \\
    & -MUL & 48.36 & 49.45 & 48.90 \\
    & -MUL + UL & 48.40 & 49.13 & 48.76 \\
    & -MUL -ME + UL & 47.99 & 49.05 & 48.52 \\
    & -MC dropout & 48.54 & 49.96 & 49.24 \\
    \midrule
    \multirow{6}{*}{$\mathtt{Rest16}$} & \textbf{Our} & \textbf{59.24} & \textbf{61.75} & \textbf{60.47} \\
    \hdashline
    & -ME & 58.81 & 60.78 & 59.77 \\
    & -MUL & 58.53 & 60.88 & 59.68 \\
    & -MUL + UL & 58.74 & 61.48 & 60.08 \\
    & -MUL -ME + UL & 58.52 & 60.80 & 59.64 \\
    & -MC dropout & 58.07 & 60.83 & 59.41 \\
    \midrule
    \multirow{6}{*}{$\mathtt{Restaurant}$} & \textbf{Our} & \textbf{61.22} & \textbf{59.87} & \textbf{60.53} \\
    \hdashline
    & -ME & 60.19 & 58.76 & 59.46 \\
    & -MUL & 60.86 & 59.45 & 60.15 \\
    & -MUL + UL & 61.00 & 59.74 & 60.36 \\
    & -MUL -ME + UL & 60.58 & 59.10 & 59.84 \\
    & -MC dropout & 59.70 & 58.36 & 59.02 \\
    \midrule
    \multirow{6}{*}{$\mathtt{Laptop}$} & \textbf{Our} & \textbf{44.38} & \textbf{43.65} & \textbf{44.01} \\
    \hdashline
    & -ME & 43.31 & 42.58 & 42.94 \\
    & -MUL & 43.87 & 43.10 & 43.48 \\
    & -MUL + UL & 42.89 & 42.03 & 42.45 \\
    & -MUL -ME + UL & 43.08 & 42.12 & 42.59 \\
    & -MC dropout & 44.19 & 43.12 & 43.65 \\
    \bottomrule
    \end{tabular}}}
    \caption{Evluation results of ablation study. The minus ``-'' denotes removing components and the addition ``+'' denotes adding components.}
    \label{table:ablation study}

\end{table}

\subsubsection{Ablation Study}

To validate the effectiveness of individual components, we perform a systematic ablation study based on Special\_Symbols+UAUL. The experimental results are presented in Table \ref{table:ablation study}. It is worth noting that -MC dropout represents removing the MC dropout and directly sampling the output of a network layer for negative samples. -MUL+UL means replacing the marginalized unlikelihood learning with the original unlikelihood. 

Firstly, it can be observed that by removing various components, the performances on four datasets are consistently decreasing. This validates the effectiveness of the constituent part of UAUL. Concretely, we see that removing MUL causes significant performance declines on all datasets. This presents that MUL is effective and telling language models \emph{what not to generate} successfully makes quadruplets extraction more accurate. Moreover, replacing MUL with naive UL also leads to performance drops. This further demonstrates that only using UL is not enough. The proposed MUL can widen the gap between correct and easily-mistaken words and is beneficial for quadruplets prediction.

Secondly, it is found that the full model slightly outperforms the variant of removing ME, suggesting that ME is able to enhance likelihood learning and balance its effects with MUL. We also observe that -MUL-ME+UL, brings consistent degradation. In most experimental settings, it is less performed than -MUL+UL. This further demonstrates the effectiveness of ME.

Finally, we also see that the full model consistently outperforms the variant of removing MC dropout on four datasets. The observation shows that understanding the uncertainty of language models is important to choose crucial mistakes. Making language models distinguish these easily-mistaken words contributes to ASQP.

\subsubsection{Comparison with Other Strategies}
\begin{figure}[t]
\centering
\includegraphics[width=0.48\textwidth]{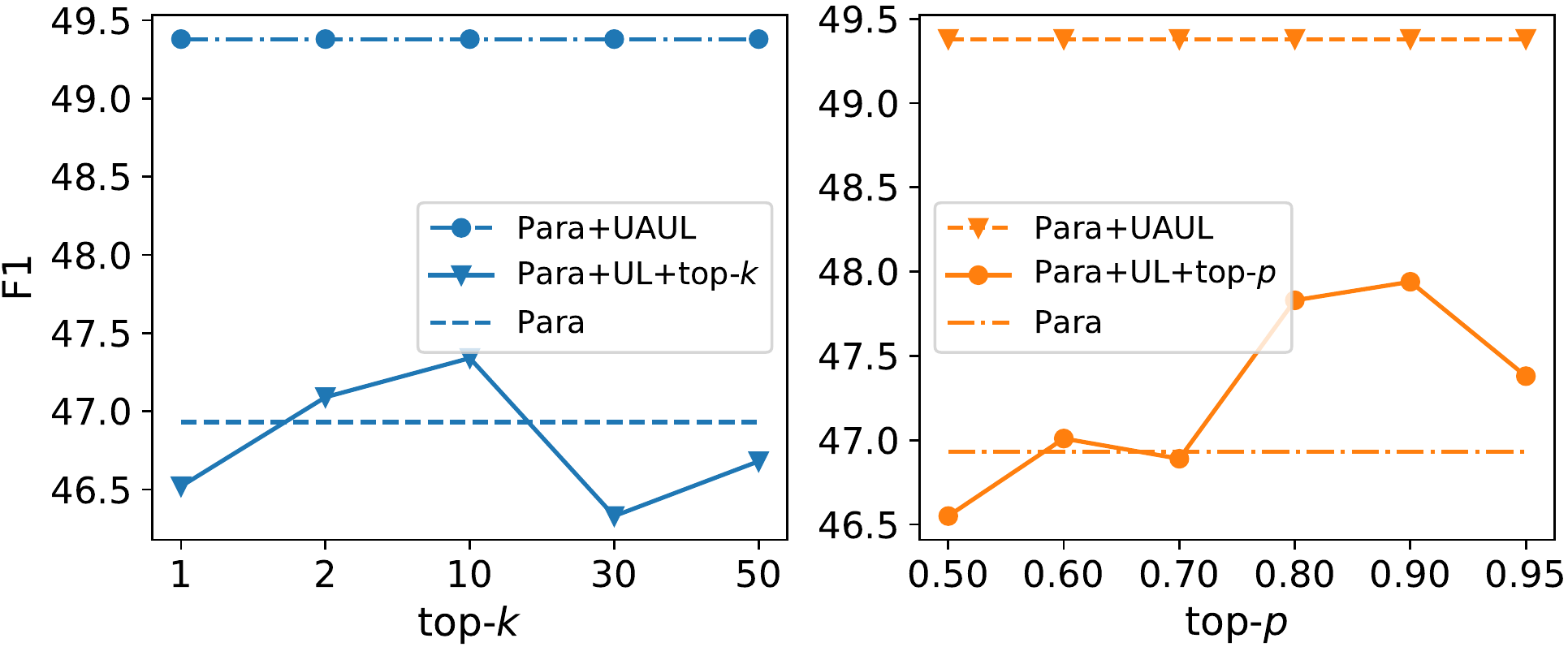}

\caption{Evaluation results of other strategies on $\mathtt{Rest15}$. Para denotes the Paraphrase method.}
\label{pic:topktopp}
\end{figure}

We further compare our method with choosing negative samples via top-$k$ \cite{fan2018hierarchical} and top-$p$ \cite{Holtzman2020The} strategies. Previously, these two strategies are exploited in the inference phase of text generation. Here we borrow their idea to select negative samples in the training phase. Specifically, except for the ground-truth token, all other samples from top-$k$ and top-$p$ are regarded as negative. The evaluation results are depicted in Figure \ref{pic:topktopp}.

We first see that introducing unlikelihood learning with top-$k$ or top-$p$ strategies can both bring some gains by setting specific $k$ or $p$ values. This demonstrates that learning negative information is effective for ASQP. Yet the gains are very limited. Then it can be observed that Paraphrase+UAUL achieves significant improvements, showing the effectiveness of our approach. This suggests that considering the uncertainty of language models can successfully choose more valuable samples.

\subsubsection{Hyperparameter Study}
\label{sec:effect of hyper}

\begin{figure}[t]
\centering
\includegraphics[width=0.47\textwidth]{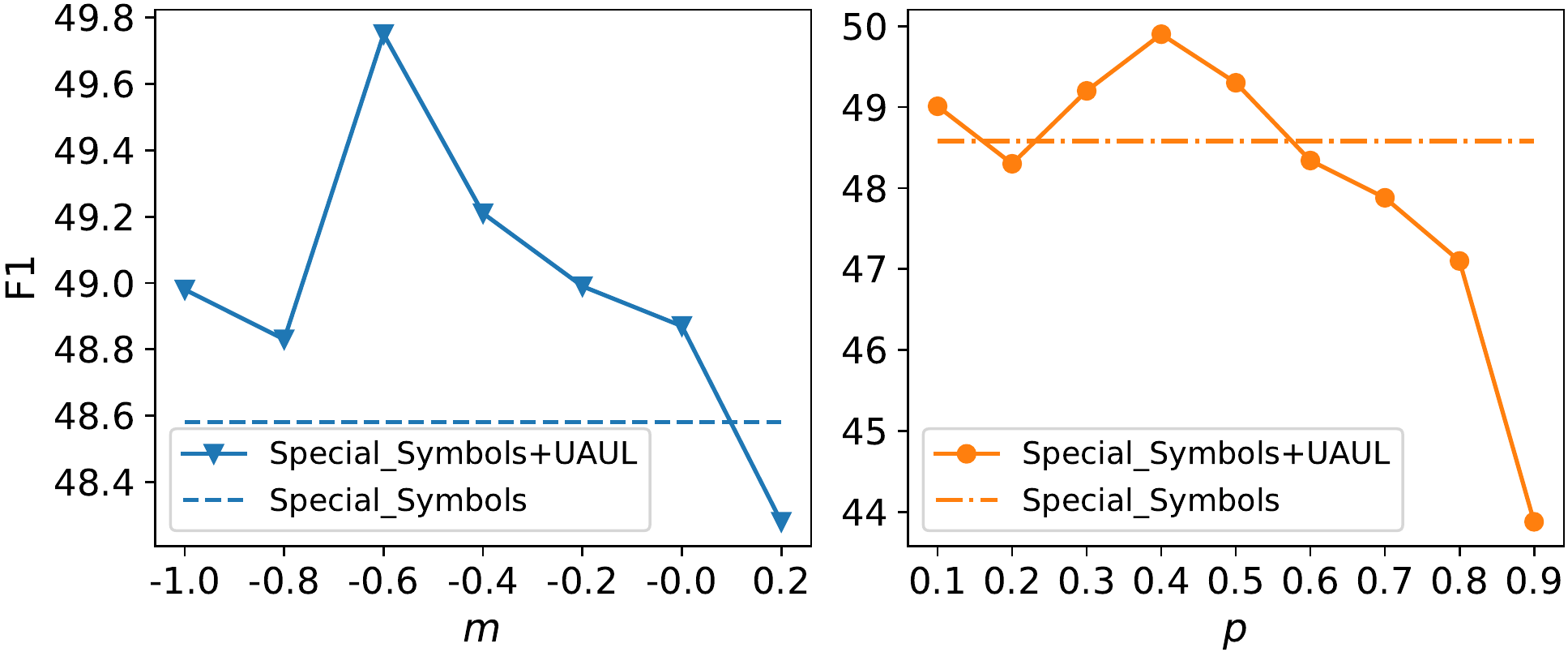}

\caption{Effect of hyperparameters on $\mathtt{Rest15}$.}
\label{pic:hyperparameters}
\end{figure}

The effects of two hyperparameters are also studied, i.e. $m$ and $p$, where $m$ is the margin in Eq. (\ref{eq:MUL}) and $p$ is the MC dropout rate. The curves are depicted in Figure \ref{pic:hyperparameters}.

\noindent
\textbf{Hyperparameter $\boldsymbol{m}$} \; It determines the gap extent to learn from negative samples. Fixing dropout to 0.4 and keeping all other parameters the same, we vary $m$ from -1.0 to 0.2. In the left plot of Figure \ref{pic:hyperparameters}, it is found that with most of the values, Special\_Symbols+UAUL outperforms the original model. It shows that this hyperparameter has robustness to some extent. Then setting $m$ too small or too large leads to a decrease in performance. If the gap extent is too large, it probably causes overfitting, while if too small, the influences of uncertainty-aware negative samples are limited.


\noindent
\textbf{Dropout Rate $\boldsymbol{p}$} \; It determines the proportions of neural connections to drop. Setting various values of $p$ will lead to different extents of uncertainty for language models. As shown in the right plot of Figure \ref{pic:hyperparameters}, we find that keeping $p$ within 0.1 to 0.5 yields good results. However, if $p$ is set to a large value, the scale of weights for optimizing is limited, which affects likelihood training and in turn, causes performance degradation.


\subsubsection{Error Analysis}
\label{sec:error}

\begin{figure}[t]
\centering
\includegraphics[width=0.48\textwidth]{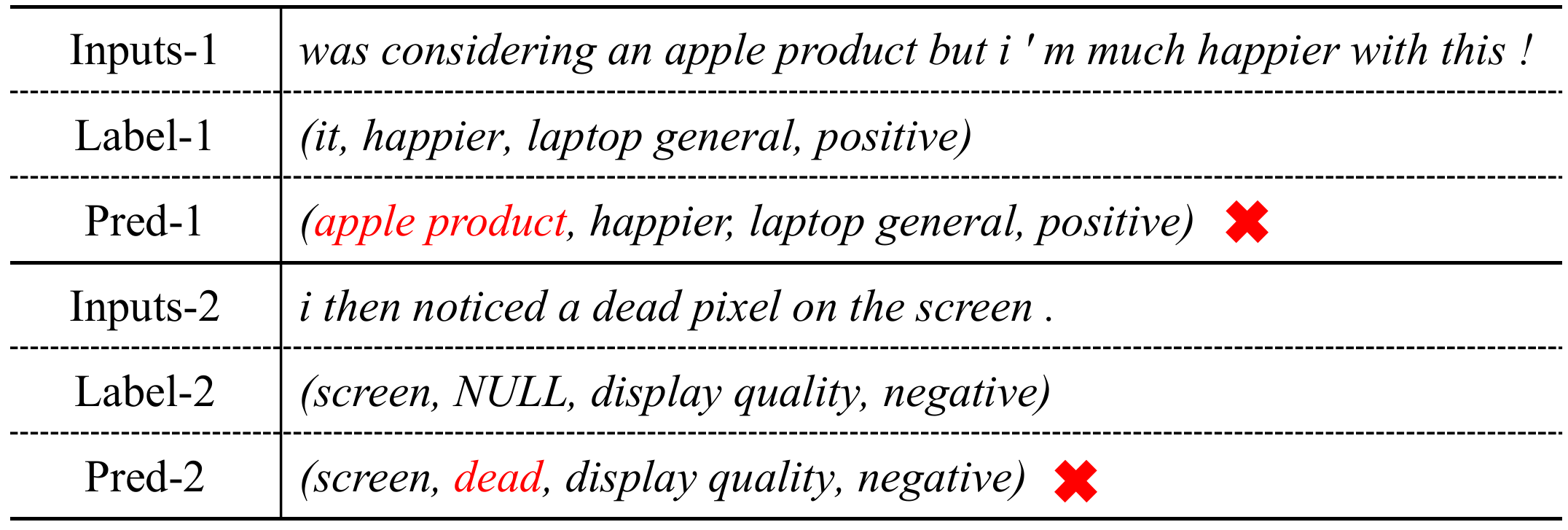} 
\caption{Two error cases predicted by Paraphrase+UAUL from the testing set of $\mathtt{Laptop}$ dataset.}
\label{fig:error_anal}
\end{figure}

To better understand the limitations of UAUL, we choose a typical method Paraphrase+UAUL, and conduct an error analysis. Two failed cases are shown in Figure \ref{fig:error_anal}. 

The first sentence implicitly describes an aspect term that the user is \emph{``much happier with''} rather than \emph{``apple product''}. Thus the ground-truth aspect term is \emph{``it''}, yet our approach predicts \emph{``apple product''}. Similarly, the second case expresses the negative opinion towards \emph{``screen''} since it has a \emph{``a dead pixel''}. The opinion term is also implicit, but our approach predicts wrongly to an adjective \emph{``dead''}. In summary, an aspect/opinion term may be described implicitly, which requires deep semantic understanding. Though UAUL achieves consistent performance improvements for various generation-based methods, it struggles to deal with implicit information.


\section{Related Work}

\textbf{Aspect-Base Sentiment Analysis} (ABSA) \; Early studies of ABSA stay at the level of individual elements, such as extracting aspect terms \cite{xu2018double}, detecting aspect categories \cite{bu2021asap,brauwers2022survey}, predicting the sentiment polarity given an aspect term \cite{huang2019parameterized} or an aspect category \cite{hu2018can}. Subsequently, researchers \cite{schouten2015survey,zhang2022survey} pay attention to the dependencies of multiple elements and recognize them simultaneously. 
\citet{peng2020knowing} focus on the triplet of aspect opinion sentiment. 
Recently, ASQP has drawn much attention, dealing with the whole elements, i.e. aspect sentiment quadruplets. To address ASQP, pipeline method \cite{cai2021aspect} and generation-based method \cite{zhang2021towards-generative,zhang2021aspect-sentiment} are proposed. Due to the simplicity and end-to-end manner, the generation paradigm has become the main research direction. Promising works design novel approaches based on tree structure \cite{mao2022seq2path,bao2022aspect}, contrastive learning \cite{peper2022generative} and data augmentation \cite{hu2022improving}. Different from the above works, we study ASQP from the perspective of what not to generate and design novel uncertainty-aware unlikelihood learning for the ASQP task.

\noindent
\textbf{Unlikelihood Learning} \; It is originally proposed in the field of neural text generation \cite{welleck2020neural}. It aims to deal with the generation repetition problem, which records the words that have been decoded and suppress their probabilities in the following decoding time steps. \citet{li2019don} introduce unlikelihood loss into dialog generation to address the utterance repetition, frequent words, and logical flaw issues. 
\newcite{song2021bob} leverage unlikelihood training to improve the understanding of character consistency in the persona-based dialogue.
In this work, semantic-similar or ambiguous tokens are negative information for ASQP. We acquire them via the inherent uncertainty of language models and propose novel marginalized unlikelihood learning to deal with negative samples.


\section{Conclusion}

Generation-based paradigm has become a new trend for ASQP. Yet previous works mainly consider what to generate but ignore what not to generate. In this work, we propose a template-agnostic uncertainty-aware unlikelihood learning (UAUL) method to address negative samples. We acquire easily-mistaken samples by modeling the built-in uncertainty of language models. Then based on the mistakes, we propose marginalized unlikelihood learning to promote the distinguishable of the noises and errors. To balance the impact of marginalized unlikelihood learning, we design minimization entropy. Extensive experiments on various generate-based methods demonstrate UAUL has universal effectiveness towards different templates.


\section*{Limitations}
Our work is the first study of generative ASQP task from the view of what not to generate. Despite the state-of-the-art performance and template-agnostic effectiveness, our work still has limitations that may guide the direction of future work.

Firstly, implicit information is still challenging for UAUL. Failed cases in error analysis \S\ref{sec:error} demonstrate that tough cases require in-depth semantic understanding. Though UAUL achieves wide improvements in the generation paradigm, it struggles to deal with implicit cases.

Secondly, in this work, we only design token-level 
marginalized unlikelihood learning. Since aspect sentiment quadruplets contain four types of information, considering span-level and whole sequence-level negative sample learning may attain further gains.


Thirdly, UAUL increases the training time, as shown in Table \ref{table:time}. We optimize the implementation by parallel computation. Meanwhile, MC dropout is only adopted in the last dropout layer. The training time is still significantly enlarged. Nevertheless, our method does not require additional human labor, which has obvious advantages in real applications.


\section*{Acknowledgements}
We sincerely thank all the anonymous reviewers for providing valuable feedback. This work is supported by the youth program of National Science Fund of Tianjin, China (Grant No. 22JCQNJC01340), the Fundamental Research Funds for the Central University, Nankai University (Grant No. 63221028 and No.63232114), the key program of National Science Fund of Tianjin, China (Grant No. 21JCZDJC00130).

\bibliography{anthology,custom}
\bibliographystyle{acl_natbib}

\appendix
\section{Experimental Details}

\subsection{Software and Hardware}
The details of the software and hardware environments are as follows.

\begin{itemize}
    \item \textbf{System}: Ubuntu 9.4.0; Python3.8; PyTorch 1.7.0
    \item \textbf{CPU}: 12 vCPU Intel(R) Xeon(R) Platinum 8255C CPU @ 2.50GHz
    \item \textbf{GPU}: NVIDIA GeForce RTX 3090
\end{itemize}

\subsection{Datasets Details}

All four of our datasets are publicly available and are linked as follows:

\begin{itemize}
    \item $\mathtt{Rest15}, \mathtt{Rest16}$: \href{https://github.com/IsakZhang/ABSA-QUAD}{https://github.com/IsakZhang/ABSA-QUAD}

    \item $\mathtt{Restaurant}, \mathtt{Laptop}$: \href{https://github.com/NUSTM/ACOS}{https://github.com/NUSTM/ACOS}
\end{itemize}

\begin{table}[]
\small
    \centering
    
    \setlength{\tabcolsep}{2.0mm}{
    \begin{tabular}{c c c}
    \toprule
    \textbf{Models} & \textbf{Learning Rate} & \textbf{Beam Size} \\
    \midrule
    GAS & 3e-4 & 1 \\
    \midrule
    Paraphrase & 3e-4 & 1 \\
    \midrule
    Special\_Symbols & 3e-4 & 1 \\
    \midrule
    DLO & 1e-4 & 5 \\
    \midrule
    ILO & 1e-4 & 5 \\
    \bottomrule
    \end{tabular}}
    \caption{Hyperparameters of baseline methods. Beam size is the number of paths searched by the beam search at the inference stage. And beam size is 1, indicating that the inference stage uses the greedy search for decoding.}
    \label{table:baseline}
    
\end{table}

\subsection{Implementations of Baseline Methods}

We choose the following open-source generation-based methods to evaluate the proposed UAUL. 
\begin{itemize}
    \item \textbf{GAS}: \href{https://github.com/IsakZhang/Generative-ABSA}{https://github.com/IsakZhang/Generative-ABSA}

    \item \textbf{Paraphrase}: \href{https://github.com/IsakZhang/ABSA-QUAD}{https://github.com/IsakZhang/ABSA-QUAD}

    \item \textbf{Special\_Symbols}, \textbf{DLO}, \textbf{ILO}: \href{https://github.com/hmt2014/AspectQuad}{https://github.com/hmt2014/AspectQuad}
\end{itemize}

\begin{figure}[t]
\centering
\includegraphics[width=0.48\textwidth]{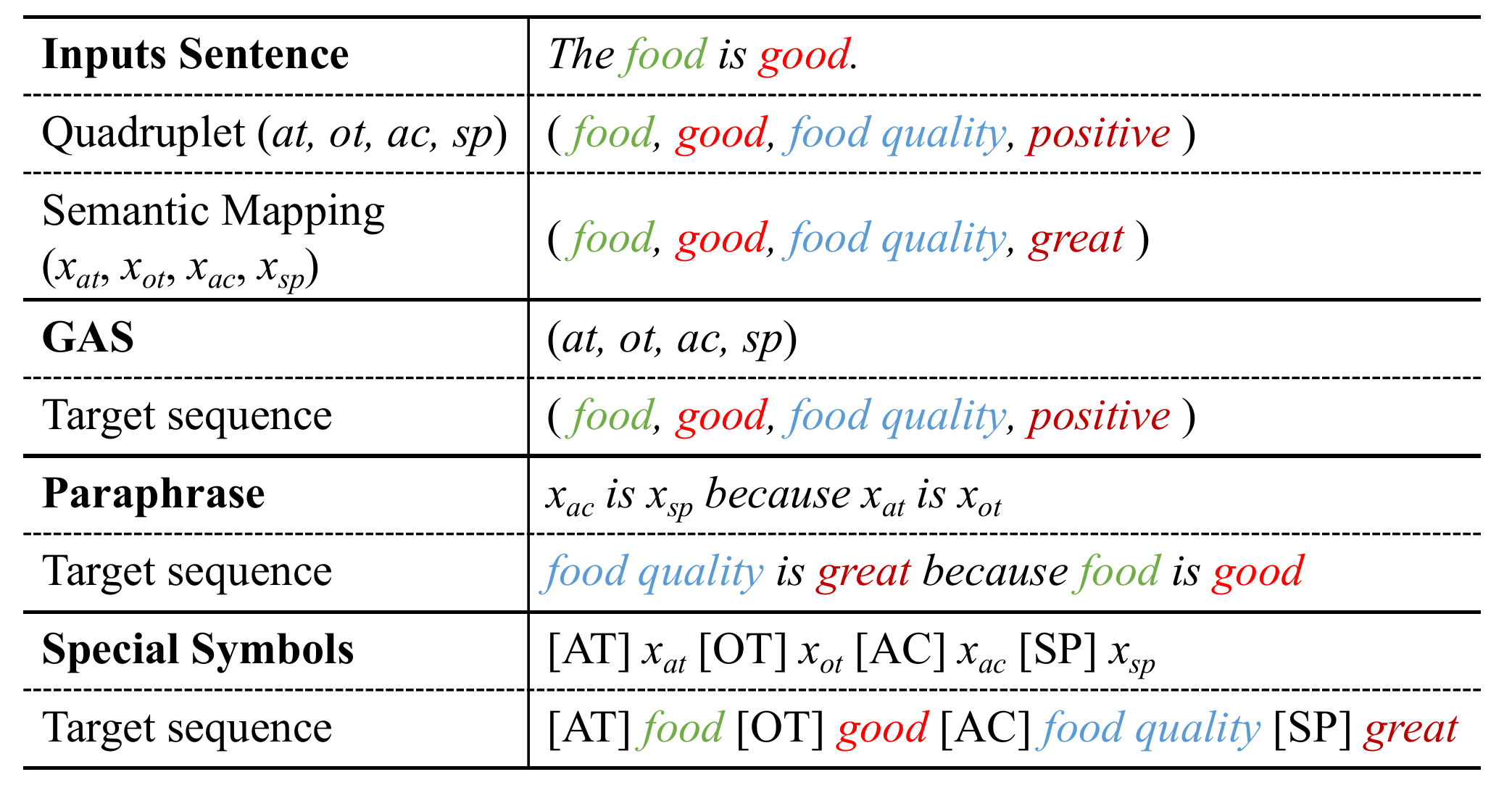} 
\caption{Template details of the various methods.}
\label{fig:template}
\end{figure}

For all baseline methods, the number of epochs is set to 20. The batch size is 16. Other parameters are shown in Table \ref{table:baseline}. In addition, we also depict the template details of each baseline method in Figure \ref{fig:template}. It is worth noting that ILO and DLO also follow the special symbols templates but combine multiple template orders as data augmentation.

\begin{table}[]
\small
    \centering
    
    \setlength{\tabcolsep}{2.0mm}{
    \begin{tabular}{c c c c c}
    \toprule
    \textbf{Datasets} & \textbf{$m$} & \textbf{$p$} & \textbf{$\alpha$} & MC forward num \\
    \midrule
    $\mathtt{Rest15}$ & -0.6 & 0.4 & 10 & 5 \\
    $\mathtt{Rest16}$ & -0.3 & 0.4 & 10 & 5 \\
    $\mathtt{Restaurant}$ & -0.6 & 0.4 & 10 & 5 \\
    $\mathtt{Laptop}$ & -0.1 & 0.4 & 10 & 5 \\
    \bottomrule
    \end{tabular}}
    \caption{Hyperparameters of the proposed UAUL.}
    \label{table:m}
\end{table}



\subsection{Implementations of Our Method \label{mainsetups}}


In the experiments, all the reported results are the average of 5 runs. 
For all baselines and our methods, we use T5-base \cite{JMLR:v21:20-074} as our pre-trained language model. And when applying our method, we keep all the parameters the same as in the baseline method \ref{table:baseline}. 
The hyperparameter details are presented in Table \ref{table:m}. 







\begin{figure}[t]
\centering
\includegraphics[width=0.48\textwidth]{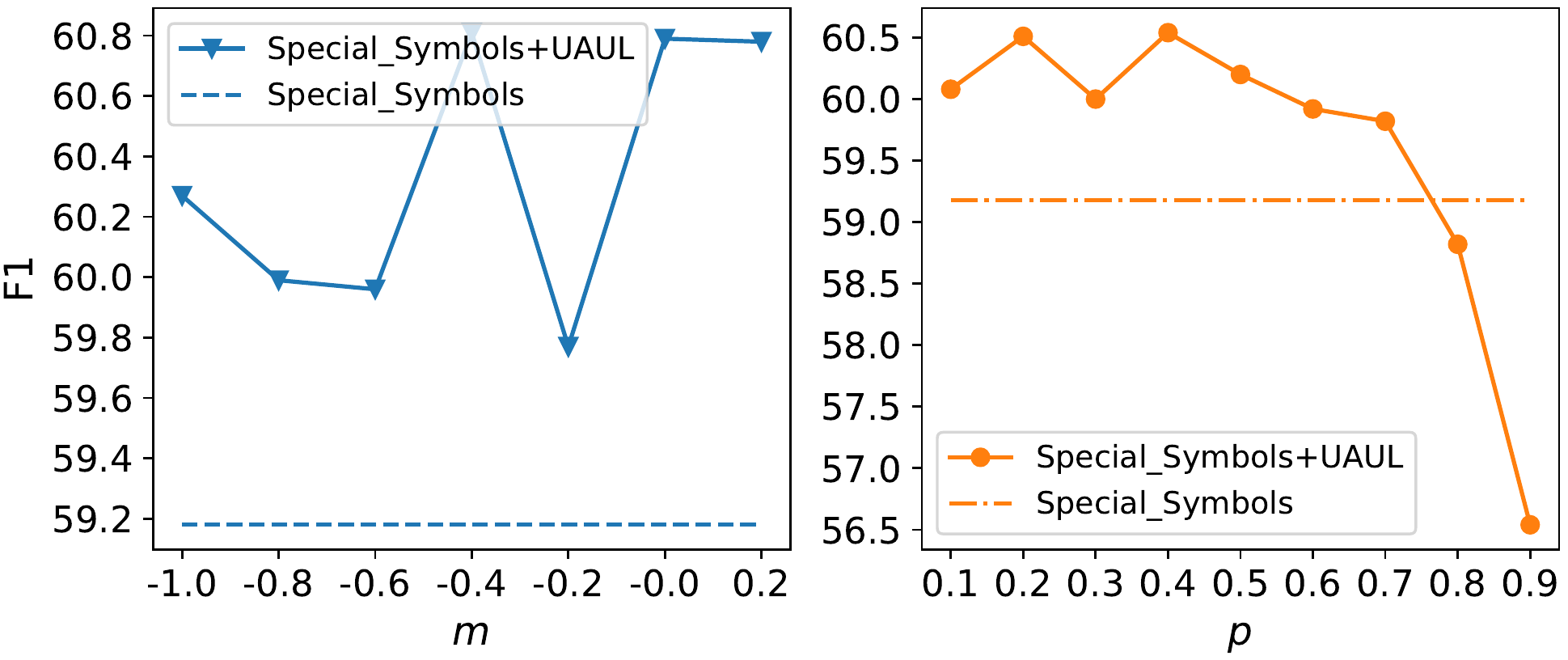}
\caption{Effect of hyperparameters on $\mathtt{Restaurant}$}
\label{pic:m-p-Rest}
\end{figure}

\begin{figure}[t]
\centering
\includegraphics[width=0.48\textwidth]{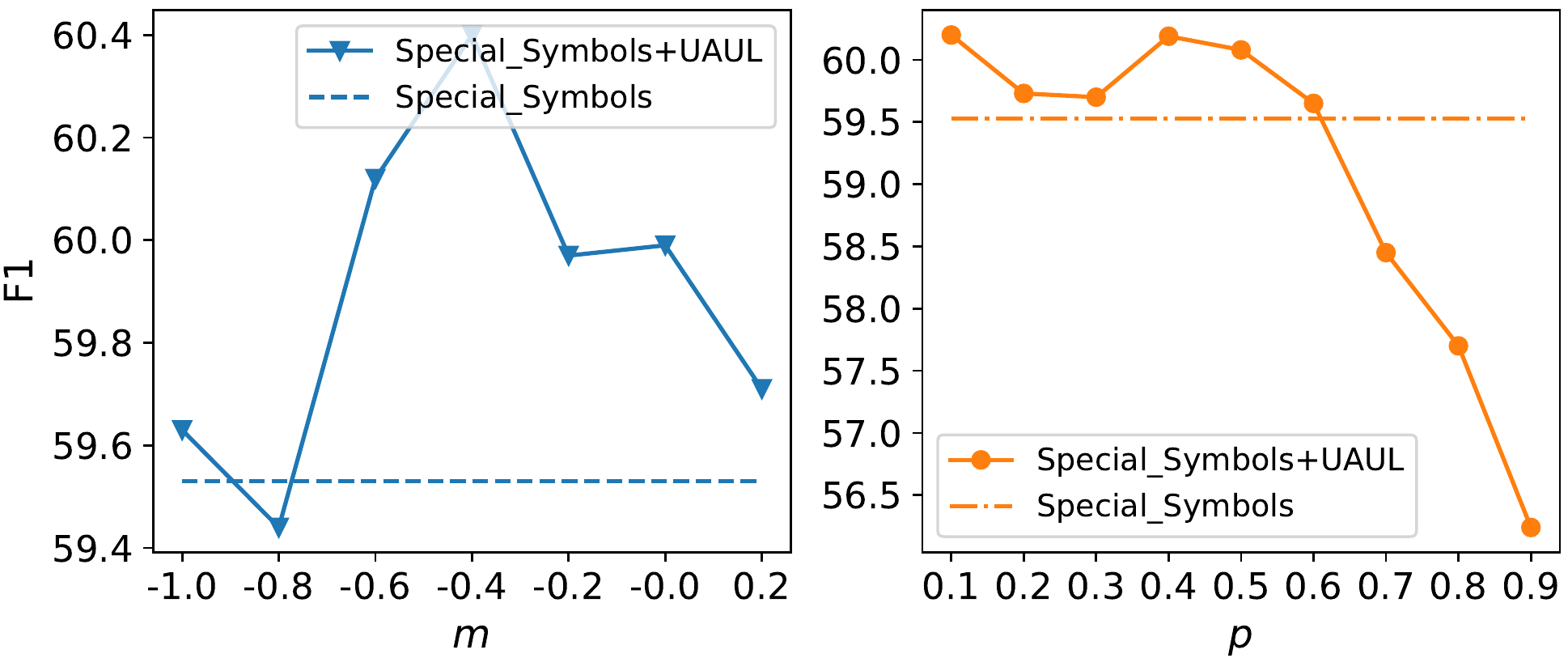}
\caption{Effect of hyperparameters on $\mathtt{Rest16}$}
\label{pic:m-p-rest16}
\end{figure}

\section{Additional Experimental Results}

\subsection{Additional Hyperparameter Study}

Hyperparameter studies on $\mathtt{Restaurant}$ and $\mathtt{Rest16}$ are depicted in Figure \ref{pic:m-p-Rest} and Figure \ref{pic:m-p-rest16}. It can be found that on $\mathtt{Restaurant}$, our method outperforms Special\_Symbols on various values of $m$. On $\mathtt{Rest16}$, our method also outperforms Special\_Symbols in most cases. This demonstrates that hyperparameter $m$ has high robustness. For the dropout rate $p$, it is found that keeping $p$ within 0.1 to 0.5, the results are the best, but setting $p$ larger than 0.7 causes performance degradation. A possible explanation is that dropping out too much scale of neural connections reduces the proportion of learnable parameters.


\begin{table}[t!]
\small
    \centering
    
    \setlength{\tabcolsep}{2.0mm}{
    \begin{tabular}{c c c c}
    \toprule
    \textbf{Models} & \textbf{$\mathtt{Rest15}$} & \textbf{$\mathtt{Restaurant}$} \\
    \midrule
    GAS & 356s & 556s \\
    +UAUL & 616s & 1043s \\
    \midrule
    Paraphrase & 347s & 555s \\
    +UAUL & 609s & 1034s \\
    \midrule
    Special\_Symbols & 316s & 623s \\
    +UAUL & 605s & 1204s \\
    \midrule
    DLO & 1400s & 2346s \\
    +UAUL & 1895s & 3516s \\
    \midrule
    ILO & 1352s & 2325s \\
    +UAUL & 1902s & 3640s \\
    \bottomrule
    \end{tabular}}
    \caption{Average running time of each model.}
    \label{table:time}
    
\end{table}

\subsection{Training Time Analysis}
The average running time of each model is shown in Table \ref{table:time}. We can observe that on five generation-based methods, UAUL consistently causes more training time. Even UAUL has already used parallel computation and the last layer MC dropout in the training phase, the training time is still extremely enlarged. Admittedly the time overhead is a limitation of our approach, but our method does not require additional human labor, which is also very beneficial in practical applications.
\end{document}